\pdfoutput=1

\documentclass[11pt]{article}

\usepackage{acl}

\usepackage{times}
\usepackage{latexsym}

\usepackage[T1]{fontenc}
\usepackage{times}
\usepackage{latexsym}
\usepackage{booktabs}
\usepackage{tabularx}
\usepackage[textsize=scriptsize]{todonotes}
\usepackage{multirow}
\usepackage{url}
\usepackage{amsmath}
\usepackage{mathtools}
\usepackage{amssymb}
\usepackage{amsthm}
\usepackage{tablefootnote}
\usepackage{caption}
\usepackage{subcaption}
\usepackage{bbm}
\usepackage{comment}
\usepackage{xspace}
\usepackage{algorithm}
\usepackage{proof}
\usepackage{stmaryrd}
\usepackage{bm}
\usepackage{pifont}

\usepackage[utf8]{inputenc}

\usepackage{microtype}

\newcommand{\STAB}[1]{\begin{tabular}{@{}c@{}}#1\end{tabular}}

\newcommand{\lime}{\textsc{Lime}}
\newcommand{\shap}{\textsc{Shap}}

\newcommand{\roberta}{\textsc{RoBERTa}}

\newcommand{\squad}{\textsc{Squad}}
\newcommand{\sqadv}{\textsc{Squad-Adv}}
\newcommand{\hotpot}{\textsc{HotpotQA}}
\newcommand{\trivia}{\textsc{TriviaQA}}

\newcommand{\mnli}{\textsc{MNLI}}
\newcommand{\qnli}{\textsc{QNLI}}
\newcommand{\mrpc}{\textsc{MRPC}}

\newcommand{\limecal}{\textsc{LimeCal}}
\newcommand{\shapcal}{\textsc{ShapCal}}
\newcommand{\bowprop}{\textsc{BowProp}}

\usepackage{xcolor,colortbl}
\definecolor{badcolor}{rgb}{0.7,1,1}
\definecolor{goodcolor}{rgb}{0.867,0.925,0.984}
\newcommand{\goodcell}{\cellcolor{goodcolor}}  

\title{Can Explanations Be Useful for Calibrating Black Box Models?}

\author{Xi Ye \quad\quad Greg Durrett\\
  Department of Computer Science \\
  The University of Texas at Austin \\
  \texttt{\{xiye,gdurrett\}@cs.utexas.edu} \\ }

\begin{document}
\maketitle

\begin{abstract}
NLP practitioners often want to take existing trained models and apply them to data from new domains. While fine-tuning or few-shot learning can be used to adapt a base model, there is no single recipe for making these techniques work; moreover, one may not have access to the original model weights if it is deployed as a black box.
We study how to improve a black box model's performance on a new domain by leveraging \emph{explanations} of the model's behavior. Our approach first extracts a set of features combining human intuition about the task with model attributions generated by black box interpretation techniques, then uses a simple calibrator, in the form of a classifier, to predict whether the base model was correct or not. We experiment with our method on two tasks, extractive question answering and natural language inference, covering adaptation from several pairs of domains with limited target-domain data. The experimental results across all the domain pairs show that explanations are useful for calibrating these models, boosting accuracy when predictions do not have to be returned on every example. We further show that the calibration model transfers to some extent between tasks.\footnote{Code available: \url{https://github.com/xiye17/InterpCalib}}
\end{abstract}

\begin{figure}[t!]
\centering
\includegraphics[width=\linewidth,trim=370 25 370 25,clip]{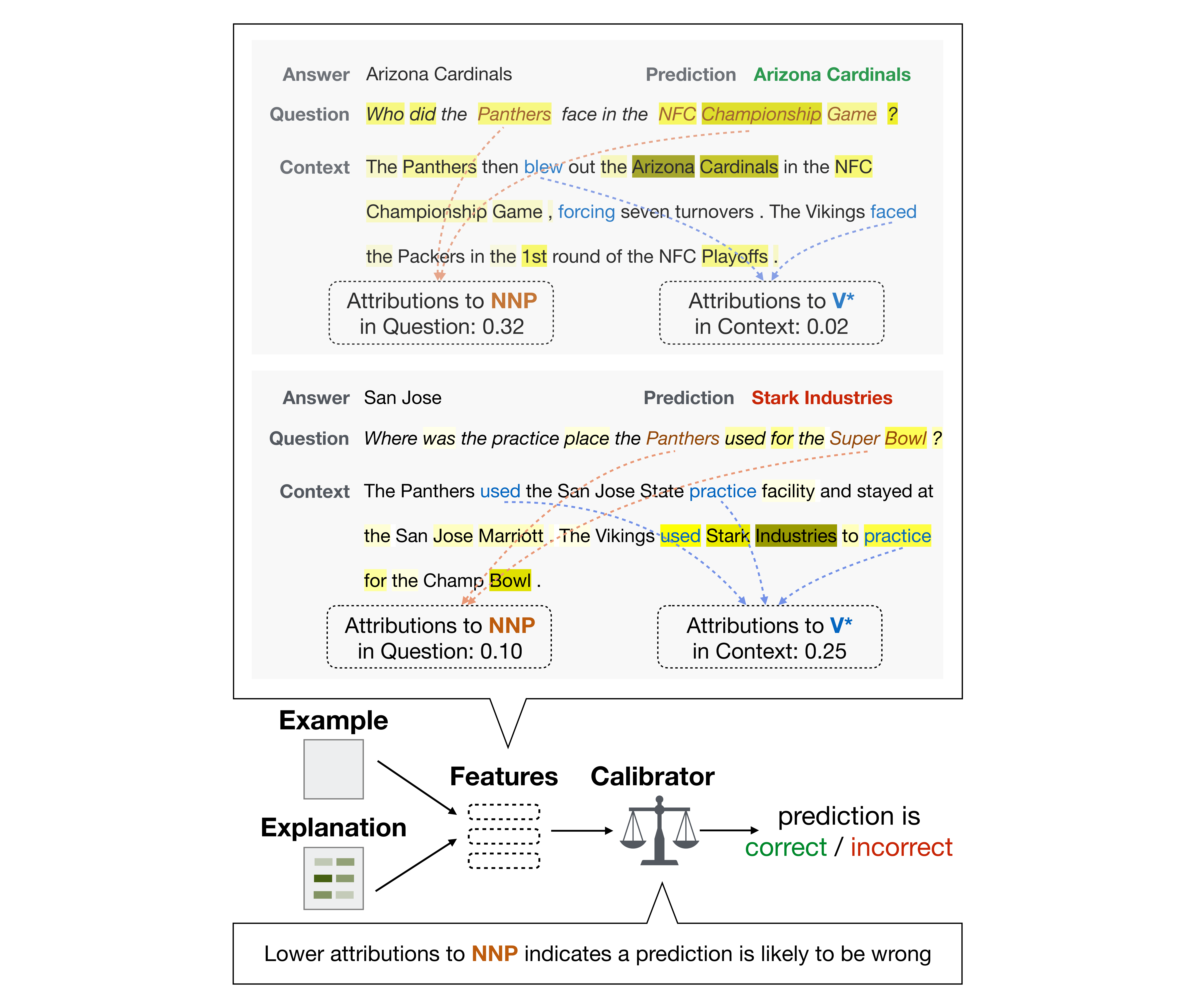}
\caption{Calibrator pipeline and examples from the \sqadv{} dataset. A \roberta{} model trained on \squad{} is correct on the first example but incorrect on the second. Features that inspect attribution values produced by \lime{} can differentiate these two on the basis of attributions to \texttt{\small NNP} in the question and \texttt{\small V*} in the context. A calibrator using these features can predict whether the original model was right or wrong.}
\label{fig:framework}
\end{figure}

\section{Introduction}
With recent breakthroughs in pre-training, NLP models are showing increasingly promising performance on real-world tasks, leading to their deployment at scale for translation, sentiment analysis, and question answering. These models are sometimes used as black boxes, especially if they are only available as a service through APIs\footnote{Google Translate, the Perspective API \url{https://perspectiveapi.com/}, and MonkeyLearn \url{https://monkeylearn.com/monkeylearn-api/} being three examples.}
or if end users do not have the resources to fine-tune the models themselves. This black-box nature poses a challenge when users try to deploy models on a new domain that diverges from the training domain, usually resulting in performance degradation.

We investigate the task of domain adaptation of black box models: given a black box model and a small number of examples from a new domain, how can we improve the model's generalization performance on the new domain? In this setting, note that we are not able to update the model parameters, which makes transfer and few-shot learning techniques inapplicable. However, we can still make the model more effective in practice by learning a \emph{calibrator}, or a separate model to make a binary decision of whether the black box model is likely to be correct or not on a given instance. While not fully addressing the domain adaptation problem, calibrating the model can make it more useful in practice, as we can recognize when it is likely to make mistakes \cite{guocalib,selectiveqa,desai2020} and modify our deployment strategy accordingly.

This paper explores how explanations can help address this task. We leverage black box feature attribution techniques \cite{lime,shap} to identify key input features the model is leveraging, even without access to model internal representations. As shown in Figure~\ref{fig:framework}, we perform calibration by connecting model interpretations with hand-crafted heuristics to extract a set of features describing the ``reasoning'' of the model. For the question answering setting depicted in the figure, answers turn out to be more reliable when the tokens of a particular set of tags (e.g., proper nouns) in the question are strongly considered. We extract a set of features describing the attribution values of different tags. Using a small number of examples in the target domain, we can train a simple calibrator for the black box model.


Our approach is closely related to the recent line of work on model behavior and explanations. \citet{chandrasekaran2018, evalai} shows explanations can help users predict model decisions in some ways and \citet{evalexplqa} show how these explanations can be semi-automatically connected to model behavior based on manually crafted heuristics. Our approach goes further by using a model to learn these heuristics, instead of handcrafting them or having a human inspect the explanations.

We test whether our method can improve model generalization performance on two tasks: extractive question answering (QA) and natural language inference (NLI). We construct generalization settings for 5 pairs of source and target domains across the two tasks. 
Compared to existing baselines \cite{selectiveqa} and our own ablations, we find explanations are indeed helpful for this task, successfully improving calibrator performance among all pairs. We even find settings where explanation-based calibrators outperform fine-tuning the model on target domain data, which assumes glass-box access to the model's parameters. Our analysis further demonstrates generalization of the calibrator models themselves: our calibrator trained on one domain can transfer to another new domain in some cases. Moreover, our calibrator can also substantially improves model performance in the Selective QA setting.



\section{Using Explanations for Black Box Model Calibration}

Let $x=x_1,x_2,...,x_n$ be a set of input tokens and $\hat{y} = f(x)$ be a prediction from our black box model under consideration. Our task in calibration\footnote{We follow \citet{selectiveqa} in treating calibration as a binary classification task. Devising a good classifier is connected to the goal of accurate estimation of posterior probabilities that calibration has more historically referred to \cite{guocalib}, but our evaluation focuses on binary accuracy rather than real-valued probabilities.} is to \textbf{assess whether the model prediction on $x$ matches its ground truth} $y$. We represent this with the variable $t$, i.e., $t\triangleq \mathbbm{1}\{f(x)= y\}$.

We explore various calibrator models to perform this task, with our main focus being on calibrator models that leverage explanations in the form of \emph{feature attribution}. Specifically, an explanation $\phi$ for the input $x$ assigns an attribution score $\phi_i$ for each input token $x_i$, which represents the importance of that token. Next, we extract features $u(x,\phi)$ depending on the input and explanation, and use the features to learn a calibrator $c:u(x,\phi)\rightarrow t$ for predicting whether a prediction is valid. We compare against baselines that do not use explanations in order to answer the core question posed by our paper's title. 

Our evaluation focuses on binary calibration, or classifying whether a model's initial prediction is correct. Following recent work in this setting \citep{selectiveqa}, we particularly focus on domain transfer settings where models make frequent mistakes. A good calibrator can identify instances where the model has likely made a mistake, so we can return a null response to the user instead of an incorrect one. 

In the remainder of this section, we'll first 
introduce how we generate the explanations and then how to extract the features $u$ for the input $x$.

\subsection{Generating Explanations}

\label{sec:expl_methods}
Since we are calibrating black box models, we adopt \lime{} \cite{lime} and \shap{} \cite{shap} for generating explanations for models instead of other techniques that require access to the model details (e.g., integrated gradients \cite{intgrad}).

The rest of this work only relies on \lime{} and \shap{} to map an input sequence $x$ and a model prediction $y$ to a set of importance weights $\phi$. We will briefly summarize the unified framework shared by both methods, and refer readers to the respective papers for additional details. 

\lime{} and \shap{} generate \emph{local explanations} by approximating the model's predictions on a set of perturbations around the base data point $x$.
In this setting, a perturbation $x'$ with respect to $x$ is a simplified input where some of the input tokens are absent (replaced with a \texttt{\small <mask>} token). Let $z=z_1,z_2,...,z_n$ be a binary vector with each $z_i$ indicating whether $x_i$ is present (using value 1) or absent (using value 0), and $h_x(z)$ be the function that maps $z$ back to the simplified input $x'$. Both methods seek to learn a local linear classifier $g$ on $z$ which matches the prediction of original model $f$ by minimizing:

\vspace{-1.5em}
\begin{gather*}
    g(z)=\phi_0 + \sum_{i=1}^{n} \phi_i z_i \\
    \xi = \mathrm{arg}\min_{g} \sum_{z\in Z} \pi_x(z) [f(h_x(z)) - g(z) ]^2 +\Omega(g) \\
\end{gather*}

where $\pi_x$ is a local kernel assigning weight to each perturbation $z$, and $\Omega$ is the L2 regularizer over the model complexity. The learned feature weight $\phi_i$ for each $z_i$ then represents the additive attribution \cite{shap} of each individual token $x_i$. \lime{} and \shap{} differ in the choice of the local kernel $\pi_x$. Please refer to the supplementary materials for details of the kernel.


\subsection{Extracting Features by Combining Explanations and Heuristics}

Armed with these explanations, we now wish to connect the explanations to \emph{the reasoning we expect from the task}: if the model is behaving as we expect, it may be better calibrated. A human might look at the attributions of some important features and decide whether the model is trustworthy in a similar fashion \cite{doshi2017towards}. Past work has explored such a technique to compare explanation techniques \cite{evalexplqa}, or run studies with human users on this task \cite{chandrasekaran2018,evalai}.

Our method automates this process by learning what properties of explanations are important. We first assign each token $x_i$ with one or more human-understandable \textbf{properties} $V(x_i)=\{v_j\}_{j=1}^{m_{i}}$. Each property $v_j \in \mathcal{V}$ is an element in the property space, which includes indicators like POS tags and is used to describe an aspect of $x_i$ whose importance might correlate with the model's robustness. We conjoin these properties with aspects of the explanation to render our calibration judgment. Figure~\ref{fig:framework} shows examples of properties such as whether a token is a proper noun (NNP).

We now construct the feature set for the prediction made on $x$. For every property $v\in \mathcal{V}$, we extract a single feature $F(v,x,\phi)$ by aggregating the attributions of the tokens associated with $v$:
$$F(v,x,\phi)=\sum_{i=1}^{n} \sum_{\bar{v} \in V(x_i)}\mathbbm{1}\{\bar{v}=v\}\phi_i $$
where $\mathbbm{1}$ is the indicator function, and $\phi_i$ is the attribution value.
An individual feature represents the total attribution with respect to property $v$ when the model is making the predictions for $x$. The complete feature set $u$ for $x$, given as $u=\{F(v,x,\phi)\}_{v\in \mathcal{V}}$, summarizes model rationales from the perspective of the properties in $\mathcal{V}$.

\paragraph{Properties} We use several types of heuristic properties for calibrating QA and NLI models.

{\bf Segments of the Input (QA and NLI):}
In both of our tasks, an input sequence can naturally be decomposed into two parts, namely a question and a context (QA) or a premise and a hypothesis (NLI). We assign each token with the corresponding segment name, which yields features like \texttt{\small Attributions to Question}.

{\bf POS Tags  (QA and NLI):}
We use tags from the English Penn Treebank \cite{marcus-etal-1993-building} to implement a group of properties. We hypothesize that tokens of some specific tags should be more important, like proper nouns in the questions of the QA tasks. If a model fails to consider proper nouns of a QA pair, it is more likely to make incorrect predictions.

{\bf Overlapping Words (NLI):}
Word overlap between a premise and a hypothesis strongly affects neural models' predictions \cite{hans}. We assign each token with the \texttt{\small Overlapping} property if a token appears in both the premise and the hypothesis, or \texttt{\small Non-Overlapping} otherwise.

{\bf Conjunction of Groups:} 
We can further produce higher-level properties by taking the Cartesian product of two or more groups.
We conjoin \texttt{\small Segment} and \texttt{\small Pos-Tags}, which yields higher-level features like \texttt{\small Attributions to NNP in Question}. Such a feature aggregates attributions of tokens that are tagged with \texttt{\small NNP} and also required to be in the question (marked with orange).



\subsection{Calibrator Model} We train the calibrator on a small number of samples in our target domain. Each sample is labeled using the prediction of the original model compared to the ground truth. Using our feature set $F(v,x,\phi)$, we learn a random forest classifier, shown to be effective for a similar data-limited setting in \citet{selectiveqa}, to predict $t$ (whether the prediction is correct). This classifier returns a score, which overrides the model's original confidence score with respect to that prediction. 

In Section~\ref{sec:experiments}, we discuss several baselines for our approach. As we vary the features used by the model, all the other details of the classifier and setup remain the same.

\section{Tasks and Datasets}
\label{sec:tasks}

\begin{figure}[t]
\centering
\includegraphics[width=\linewidth,trim=100 130 100 130,clip]{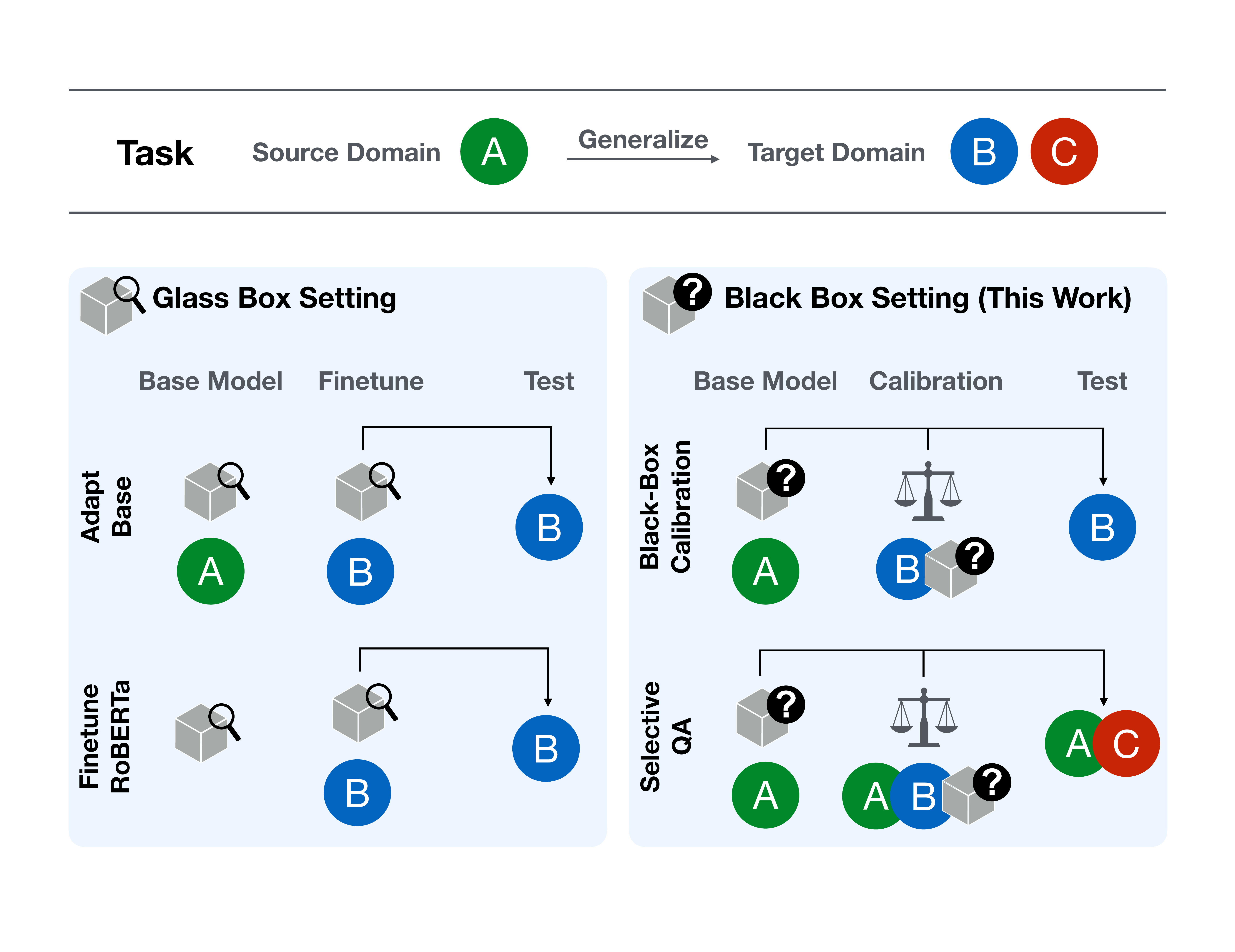}
\caption{Illustration of different settings in the experiments. In black box settings, a calibrator is trained for improving model performance on OOD data; in glass box settings, the model is finetuned on OOD data from a base model or vanilla \roberta{} LM model.}
\label{fig:setting}
\end{figure}

Our task setup involves transferring from a source domain/task A to a target domain/task B. Figure~\ref{fig:setting} shows the data conditions we operate in. Our primary experiments focus on using our features to either calibrate or selectively answer in the black box setting (right side in Figure~\ref{fig:setting}). In this setting, we have a black box model trained on a source domain A and a small amount of data from the target domain B. Our task is to train a calibrator using data from domain B to identify instances where the model potentially fails in the large unseen test data in domain B. We contrast this black box setting with glass box settings (left side in Figure~\ref{fig:setting}), where we directly have access to the model parameters and can fine-tune on domain B or train on B from scratch.



\paragraph{English Question Answering}
We experiment with domain transfer from \squad{} \cite{squaddataset} to three different settings: \sqadv{} \cite{squadadv}, \hotpot{} \cite{hotpot}, and \trivia{} \cite{triviaqa}.

\sqadv{} is an adversarial setting based on \squad{}, which constructs adversarial QA examples based on \squad{} by appending a distractor sentence at the end of each example's context. The added sentence contains a spurious answer and usually has high surface overlapping with the question so as to fool the model. We use the {\sc AddSent} setting from \citet{squadadv}.

Similar to \squad{}, \hotpot{} also contains passages extracted from Wikipedia, but \hotpot{} asks questions requiring multiple reasoning steps, although not all questions do \cite{chen-durrett-2019-understanding}.
\trivia{} is collected from Web snippets, which present a different distribution of questions and passages than \squad{}.  For  \hotpot{} and \trivia{}, we directly use the pre-processed version of dataset from the {\sc MRQA} Shared Task \cite{mrqa}.

\paragraph{English NLI}
For the task of NLI, we transfer a model trained on \mnli{} \cite{mnli} to \mrpc{} \cite{mrpc} and \qnli{} \cite{glue}, similar to the settings in \citet{madomain}. \qnli{} contains a question and context sentence pair from \squad{}, and the task is to verify whether a sentence contains the answer to the paired question. \mrpc{} is a paraphrase detection dataset presenting a binary classification task to decide whether two sentences are paraphrases of one another. Note that generalization from \mnli{} to \qnli{} or \mrpc{} not only introduces shift in terms of the distribution of the input text, but in terms of the nature of the task itself, since \qnli{} and \mrpc{} aren't strictly NLI tasks despite sharing some similarity. Both are \emph{binary} classification tasks rather than three-way.

\section{Experiments}
\label{sec:experiments}

\begin{figure}[t]
\centering
\includegraphics[width=\linewidth,trim=35 20 0 20,clip]{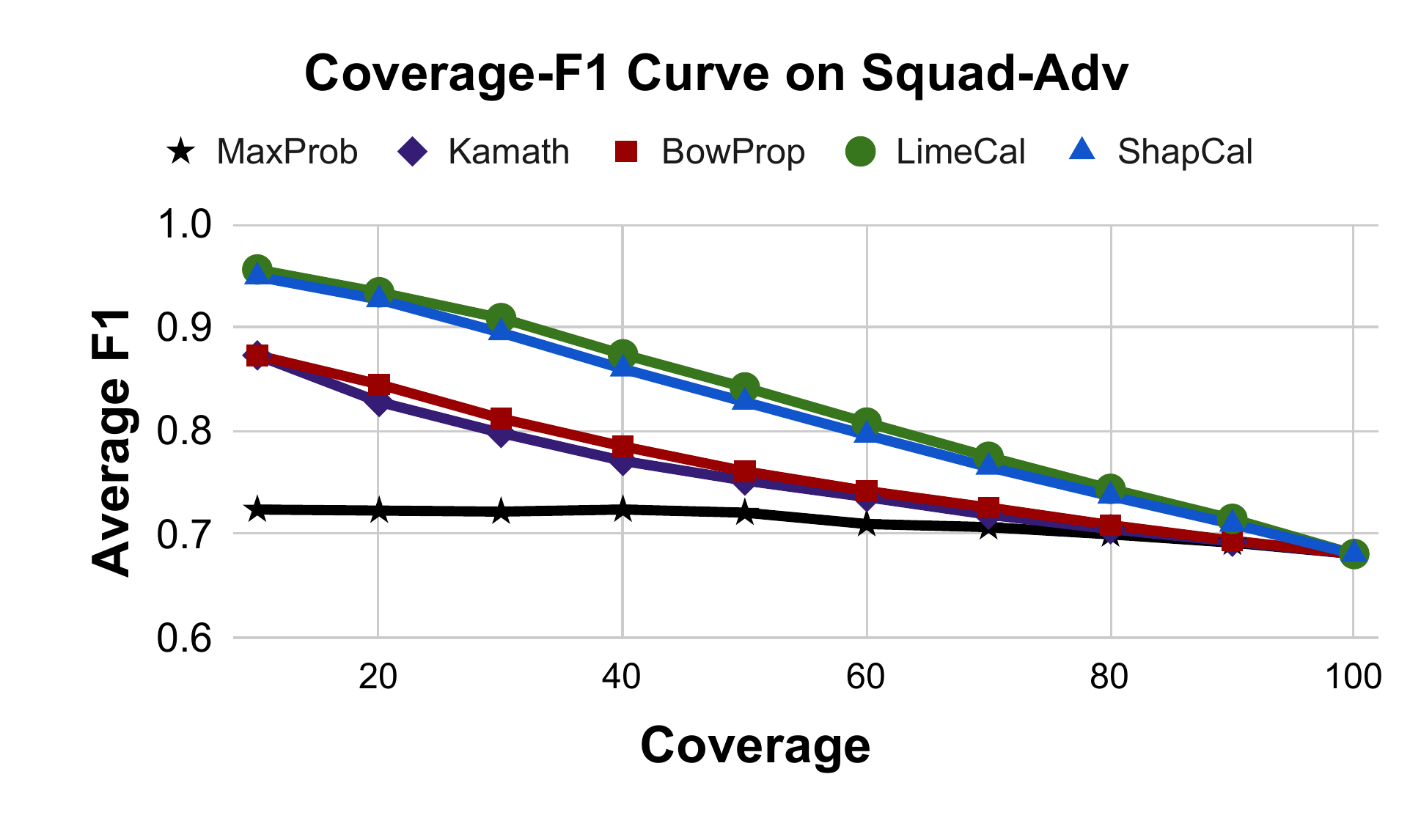}
\caption{Coverage-F1 curves of different approaches on \sqadv{}. As more low-confidence questions are answered, the average F1 scores decrease. We use AUC to evaluate calibration performance. }
\label{fig:f1auccurve}
\end{figure}

\paragraph{Baselines}
We compare our calibrator against existing baselines as well as our own ablations.

{\bf \textsc{MaxProb}} simply uses the thresholded probability of the predicted class to assess whether the prediction is trustworthy. 

{\bf \textsc{Kamath}} \cite{selectiveqa} (for QA only) is a baseline initially proposed to distinguish out-of-distribution data points from in-domain data points in the selective QA setting (see Section~\ref{sec:selective}), but it can also be applied in our settings.
It trains a random forest classifier to learn whether a model's prediction is correct based on several heuristic features, including the probabilities of the top 5 predictions, the length of the context, and the length of the predicted answer.
Since we are calibrating black box models, we do not use dropout-based features in \citet{selectiveqa}.

{\bf \textsc{ClsProbCal}} (for NLI only) uses more detailed information than {\sc MaxProb}: it uses the predicted probability for \texttt{Entailment}, \texttt{Contradiction}, and \texttt{Neutral} as the features for training a calibrator instead of only using the maximum probability.

{\bf \textsc{BowProp}} adds a set of heuristic property features on top of the {\sc Kamath} method. These are the same as the features used by the full model \emph{excluding the explanations}. We use this baseline to give a baseline for using general ``shape'' features on the inputs \emph{not} paired with explanations. 

\paragraph{Implementation of Our Method} We refer our explanation-based calibration method using explanations produced by \lime{} and \shap{} as {\bf \textsc{LimeCal}} and {\bf \textsc{ShapCal}} respectively. We note that these methods also take advantages of the bag-of-word features in \bowprop{}. For QA, the property space is the union of low-level \texttt{\small Segment} and \texttt{\small Segment} $\times$ \texttt{\small Pos-Tags}. For NLI, we use the union of \texttt{\small Segment} and \texttt{\small Segment} $\times$ \texttt{\small Pos-Tags} $\times$ \texttt{\small Overlapping Words} to label the tokens. Detailed numbers of features can be found in the Appendix.

\begin{table*}[t]
    \centering
    \footnotesize
    \begin{tabular}{l|cccccccccc}
            \toprule
          &  \multicolumn{10}{c}{\sqadv{}} \\
         Approach & Acc & $\Delta${\sc Bow} & AUC & $\Delta${\sc Bow} & F1@25 & $\Delta${\sc Bow} & F1@50 & $\Delta${\sc Bow} & F1@75 & $\Delta${\sc Bow} \\
         \midrule
         {\sc MaxProb} & 62.6& $-$& 70.9 &$-$& 72.4 &$-$& 72.1 & $-$& 70.4 & $-$\\
         {\sc Kamath} & 63.2& $-$& 76.8 & $-$& 81.4 &$-$& 75.2 & $-$& 71.2 &$-$\\
         \midrule
         {\sc BowProp} & 63.6 & 0 & 77.4 & 0& 82.9 & 0& 76.1 & 0& 71.7 &0\\
         {\sc LimeCal} & {\bf 70.3} & 6.7$\pm$1.6& {\bf 83.9} & 6.4$\pm$1.4 & {\bf 92.3} & 9.4$\pm$2.3& {\bf  84.2} & 8.1$\pm$1.6 & {\bf  75.9} & 4.2$\pm$1.0\\
         {\sc ShapCal} & 69.3 & 5.6$\pm$1.8 & 82.9 & 5.5$\pm$1.3 & 91.2 & 8.2$\pm$2.2 & 82.8 & 6.7$\pm$1.4 &  75.0 &  3.3$\pm$0.9 \\
         \midrule
          &  \multicolumn{10}{c}{\trivia{}} \\
        Approach & Acc & $\Delta${\sc Bow} & AUC & $\Delta${\sc Bow} & F1@25 & $\Delta${\sc Bow} & F1@50 & $\Delta${\sc Bow} & F1@75 & $\Delta${\sc Bow} \\
         \midrule
         {\sc MaxProb} & 67.0 &$-$ & 76.7 &$-$ & 82.1 &$-$ & 76.3 &$-$ & 71.0 &$-$\\
         {\sc Kamath} & 70.6 &$-$& 76.6 &$-$ & 82.1&$-$ & 77.9 &$-$ & 71.1&$-$ \\
         \midrule
         {\sc BowProp} & 71.2 &0  & 77.6 &0  & 84.2 &0  & 79.1 &0  & 71.6 &0   \\
         {\sc LimeCal} & {\bf 72.0} & 0.8$\pm$0.4 & {\bf  78.7} & 1.1$\pm$0.2& {\bf 85.4} &1.2$\pm$0.8 & {\bf 79.6} & 0.5$\pm$ 0.3& {\bf  72.3}  &0.8$\pm$0.2\\
         {\sc ShapCal} & 71.8  &0.6$\pm$0.4 & 78.2 &0.6$\pm$0.3& 84.7& 0.5$\pm$0.8 &79.4 &0.3$\pm$ 0.4 & 72.3 & 0.8$\pm$0.3\\
         \midrule
          &  \multicolumn{10}{c}{\hotpot{}} \\
        Approach & Acc & $\Delta${\sc Bow} & AUC & $\Delta${\sc Bow} & F1@25 & $\Delta${\sc Bow} & F1@50 & $\Delta${\sc Bow} & F1@75 & $\Delta${\sc Bow} \\
         \midrule
         {\sc MaxProb} & 63.1 & $-$ & 75.7 & $-$ & 79.7 & $-$ & 75.9 & $-$& 72.2& $-$\\
         {\sc Kamath} & 64.5& $-$ & 76.8 & $-$ & 80.8 & $-$ & 77.2 & $-$& 72.8 & $-$ \\
         \midrule
         {\sc BowProp} & 64.7 &0 & 76.6 & 0 & 80.3 & 0 & 76.9 &0 & 72.4 & 0\\
         {\sc LimeCal} & {\bf 65.7} & 1.0$\pm$0.4 & {\bf 78.2} &1.6$\pm$0.4& {\bf  82.6} &2.2$\pm$0.8& {\bf 78.4} &1.5$\pm$0.6& {\bf 73.8} &1.4$\pm$0.3\\
         {\sc ShapCal} & 65.3 &0.7$\pm$0.4& 77.8 &1.2$\pm$0.3 & 82.0 &1.6$\pm$0.7& 78.0 &1.0$\pm$0.5 & 73.5 &1.1$\pm$0.4\\
         \bottomrule
    \end{tabular}
    \caption{Main results on QA tasks. Our explanation-based methods (\limecal{} and \shapcal{}) successfully calibrate a \roberta{} QA model trained on \squad{} when transferring to three new domains, and outperform a prior approach ({\sc Kamath}) as well as our ablation using only heuristic labels (\bowprop{}). In addition, we show the mean and standard deviation of the deltas w.r.t. \bowprop{} across multiple random seeds in $\Delta${\sc Bow}.}
    \label{tab:main_qa}
\end{table*}

\subsection{Main Results: QA}

\paragraph{Setup} We train a \roberta{} \cite{roberta} QA model on \squad{} as the base model, which achieves 85.5 exact match and 92.2 F1 score. 
For the experiments on \hotpot{} and \trivia{}, we split the dev set, sample 500 examples for training, and leave the rest for testing.\footnote{Details of hyperparameters can be found in the Appendix.} For experiments on \sqadv{}, we remove the unmodified data points in the {\sc Add-Sent} setting and also use 500 examples for training. For the experiments across all pairs, we randomly generate the splits, test the methods 20 times, and average the results to alleviate the influence of randomness.

\paragraph{Metrics}
In addition to {\it calibration accuracy} ({\bf ACC}) that measures the accuracy of the calibrator, we also use the {\it area under coverage-F1 curve} ({\bf AUC}) to evaluate the calibration performance for QA tasks in particular.
The coverage-F1 curve (Figure~\ref{fig:f1auccurve}) plots the average F1 score of the model achieved when the model only chooses to answer varying fractions (coverage) of the examples ranked by the calibrator-produced confidence. A better calibrator should assign higher scores to the questions that the models are sure of, thus resulting in higher area under the curve; note that AUC of 100 is impossible since the F1 is always bounded by the base model when every question is answered. We additionally report the average scores when answering the top 25\%, 50\%, and 75\% questions, for a more intuitive comparison of the performance.

\paragraph{Results}
Table~\ref{tab:main_qa} summarizes the results for QA. First, we show that explanations are helpful for calibrating black box QA models out-of-domain. Our method using \lime{} substantially improves the calibration AUC compared to {\sc Kamath} by 7.1, 2.1 and 1.4 on \sqadv{}, \trivia{}, and \hotpot{}, respectively. In particular, \limecal{} achieves an average F1 score of 92.3 at a coverage of 25\% on \sqadv{}, close to the performance of the base model on original \squad{} examples. Our explanation-based approach is effective at identifying the examples that are robust with respect to the adversarial attacks.

Comparing \limecal{} against \bowprop{}, we find that the explanations themselves do indeed help. On \sqadv{} and \hotpot{}, \bowprop{} performs on par with or only slightly better than {\sc Kamath}. These results show that connecting explanations with annotations is a path towards building better calibrators.

Finally, we compare the performance of our methods based on different explanation techniques. \limecal{} slightly outperforms \shapcal{} in all three settings. As discussed in Section~\ref{sec:expl_methods}, \shap{} assigns high instance weights to those perturbations with few activated features. While such a choice of the kernel is effective in tasks involving tabular data \cite{shap}, this might not be appropriate for the task of QA when such perturbations may not yield meaningful examples.

\subsection{Main Results: NLI}
\paragraph{Setup} Our base NLI model is a \roberta{} classification model trained on  \mnli{}, which achieves 87.7\% accuracy on the development set. We collapse \texttt{contradiction} and \texttt{neutral} into \texttt{non-entailment} when evaluating on \qnli{} and \mrpc{}.  We continue using random forests as the calibrator model. We evaluate the generalization performance on the development sets of \qnli{} and \mrpc{}. Similar to the settings in QA, we use 500 examples to train the calibrator and test on the rest for each of the 20 random trials.

\begin{table*}[t]
    \centering
    \small
    \begin{tabular}{l|cccc|cccc}
            \toprule
           &  \multicolumn{4}{c|}{\qnli{}} & \multicolumn{4}{c}{\mrpc{}}\\
         Approach & Acc & $\Delta${\sc Bow} & AUC & $\Delta${\sc Bow} & Acc & $\Delta${\sc Bow} & AUC& $\Delta${\sc Bow}\\
         \midrule
         {\sc MaxProb} & 50.5 &$-$& 41.2 &$-$ & 57.0 &$-$ &50.0 &$-$\\
         {\sc ClsProbCal} & 56.7 &$-$ & 59.5 &$-$& 71.5 &$-$& 77.9 &$-$\\
         \midrule
         {\sc BowProp} & 74.0 &0& 82.0 &0& 71.8 &0& 79.3 &0\\
         {\sc LimeCal} & {\bf 75.0} & 1.0$\pm$0.4& {\bf 82.6} &0.7$\pm$0.4& {\bf 73.6} &1.8$\pm$1.3& {\bf 81.0} &1.7$\pm$0.9\\
         {\sc ShapCal} & 74.2&0.2$\pm$0.4& 81.9 &0.0$\pm$0.4& 73.5 &1.7$\pm$1.2 & 80.7 &1.4$\pm$0.8\\
         \bottomrule
    \end{tabular}
    \caption{Main results on NLI tasks. \limecal{} moderately improves the performance of the base \mnli{} model on \qnli{} and \mrpc{}, despite how different these tasks are from the base MNLI setting.}
    \label{tab:main_nli}
\end{table*}

\paragraph{Metrics} 
Because \qnli{} and \mrpc{} are binary classification tasks, predicting whether a model is correct (our calibration setting) is equivalent to the original prediction task. We can therefore measure calibrator performance with standard classification accuracy and AUC.

\paragraph{Results}
We show results on NLI tasks in Table~\ref{tab:main_nli}. The base {\sc MNLI} model utterly fails when transferring to \qnli{} and \mrpc{} and achieves an accuracy of 49\% and 57\%, respectively, whereas the majority class is 50\% (\qnli{}) and 65\% (\mrpc{}). With heuristic annotations, \bowprop{} is able to solve 74\% of the \qnli{} instances and 72\% of the \mrpc{} instances. Our heuristic itself is strong for \qnli{} compared to {\sc MaxProb}. \limecal{} is still the best in both settings, moderately improving accuracy by 1\% and 2\% over \bowprop{} using explanations. The results on NLI tasks suggest our method can still learn useful signals for indicating model reliability even if the underlying tasks are very different.

\subsection{Analysis}

\begin{table}[t]
    \centering
    \small
    \begin{tabular}{cl|c|c|c}
            \toprule
         \multicolumn{2}{l|}{Source\, \textbackslash \, Target}  &  {\sc Sq-Adv} &  {\sc Trivia} & {\sc Hotpot}  \\
         \midrule
         \multirow{4}{*}{\STAB{\rotatebox[origin=c]{90}{\sc Sq-Adv}}}
         & {\sc Adapt} &\multirow{4}{*}{70.9} & 76.1 & 65.8\\
         & {\sc Kamath} && 73.3 & 75.1\\
         &{\sc BowProp} & & 71.9 & 74.1\\
         & {\sc LimeCal} & & 72.9 & 71.4\\
        \midrule
         \multirow{4}{*}{\STAB{\rotatebox[origin=c]{90}{\sc Trivia}}}
         & {\sc Adapt} & 64.2 & \multirow{4}{*}{76.7} & \bf \goodcell 77.2\\
         & {\sc Kamath} & 70.5 & & \goodcell 76.7 \\
         &{\sc BowProp} & 67.1 &  & 75.0 \\
         & {\sc LimeCal} & 69.3 & & \goodcell 77.0 \\
        \midrule
         \multirow{4}{*}{\STAB{\rotatebox[origin=c]{90}{\sc Hotpot}}}
         & {\sc Adapt} & 56.6 & 74.0 &  \multirow{4}{*}{75.7}\\
         & {\sc Kamath} & 70.6 & \goodcell 77.0 & \\
         &{\sc BowProp} & 69.1 & 76.9 & \\
         & {\sc LimeCal} & 68.8 & \bf \goodcell 77.9 & \\
         \bottomrule
    \end{tabular}
    \caption{Area under Coverage-F1 curve for cross-domain calibration results. The numbers along the diagonal shows the {\sc MaxProb} performance. A better performance than {\sc MaxProb} suggests the calibrator is able to usefully generalize  (colored cells).}
    \label{tab:cross_qa}
\end{table}

\begin{table}[t]
    \centering
    \small
    \begin{tabular}{cl| c c c}
            \toprule
          \multicolumn{2}{c}{\sc QA} &  {100} &  {300} & {500}  \\
         \midrule
         \multirow{4}{*}{\STAB{\rotatebox[origin=c]{90}{\sc Sq-Adv}}}
         & {\sc MaxProb} & \multicolumn{3}{c}{70.9}\\
         & {\sc Kamath} & 72.7 & 75.6 & 76.8 \\
         &{\sc BowProp} & 75.0 & 76.0 & 77.4\\
         & {\sc LimeCal} & \bf 78.7	& \bf 82.7 & \bf 83.9 \\
        \midrule
         \multirow{4}{*}{\STAB{\rotatebox[origin=c]{90}{\sc Trivia}}}
         & {\sc MaxProb} & \multicolumn{3}{c}{76.7}\\
         & {\sc Kamath} &  74.8 & 76.2 & 76.6 \\
         &{\sc BowProp}  & 76.1 & 77.4 & 77.6 \\
         & {\sc LimeCal}  & \bf 77.2 &  \bf 78.2 & \bf 78.7 \\
        \midrule
         \multirow{4}{*}{\STAB{\rotatebox[origin=c]{90}{\sc Hotpot}}}
         & {\sc MaxProb} & \multicolumn{3}{c}{75.7}\\
         & {\sc Kamath} &  75.2 & 76.5 & 76.8 \\
         &{\sc BowProp}  & 74.9 & 76.3 & 76.6 \\
         & {\sc LimeCal}  & \bf 76.5 & \bf 77.7 & \bf 78.2 \\
        \midrule
        \multicolumn{2}{c}{\sc NLI} &  {100} &  {300} & {500}  \\
         \midrule
         \multirow{4}{*}{\STAB{\rotatebox[origin=c]{90}{\sc QNLI}}}
         & {\sc MaxProb} & \multicolumn{3}{c}{41.2}\\
         & {\sc Kamath} & 56.4 & 58.1 & 59.5 \\
         &{\sc BowProp} & 79.0 & 81.5 & 82.0\\
         & {\sc LimeCal} & \bf 79.1 & \bf 81.8 & \bf 82.8 \\
        \midrule
         \multirow{4}{*}{\STAB{\rotatebox[origin=c]{90}{\sc MRPC}}}
         & {\sc MaxProb} & \multicolumn{3}{c}{50.0}\\
         & {\sc Kamath} &  73.7	& 76.8	& 77.9 \\
         &{\sc BowProp} & 69.4	& 77.5 & 79.3 \\
         & {\sc LimeCal} & \bf 76.1 & \bf 79.9 & \bf 81.0 \\
         \bottomrule
    \end{tabular}
    \caption{AUC scores of the calibrators trained with varying training data size. Explanation-based calibrators can still learn even with limited training resources, whereas {\sc Kamath} and \bowprop{} are not effective and underperform the {\sc MaxProb} baseline on \trivia{} and \hotpot{}.}
    \label{tab:train_size}
\end{table}

\begin{table*}[t]
    \centering
    \small
    \begin{tabular}{l|cccccc|cc}
            \toprule
           &  \multicolumn{2}{c}{\sqadv{}} & \multicolumn{2}{c}{\trivia{}} & \multicolumn{2}{c|}{\hotpot{}} & {\qnli{}} & {\mrpc{}} \\
          \midrule
         \bf Model Performance  & Ex & F1 & Ex & F1 & Ex & F1 & Acc & Acc \\
         {\sc Base QA/NLI} & 62.1 & 68.0 & 53.2 & 62.1 & 50.7 & 66.3 & 50.5 & 57.2 \\
         {\sc Finetune \roberta{}} & 32.3 & 42.0 & 28.5 & 34.8 & 39.5 & 54.8 & 81.2 & 79.8\\
         {\sc Adapt Base QA/NLI} & 77.3 & 84.3 & 56.2 & 64.0 & 54.3 & 70.8 & 80.7 & 79.1 \\
         {\sc InDomain QA/NLI} & $-$ & $-$  & 62.1 & 68.1 & 59.7 & 77.2 & 92.0 & 87.2 \\
         \midrule
         \bf Calibration Results  & Acc & AUC & Acc & AUC & Acc & AUC & Acc & Acc \\
         {\sc Finetune \roberta{} + MaxProb} & $-$  & 41.1 & $-$  & 37.6 & $-$ & 67.0 & \bf 81.2 & \bf 79.8\\
         {\sc Adapt Base QA/NLI + MaxProb} & $-$  & \bf 92.7 & $-$ & 77.6 & $-$ & \bf 82.5 & 80.7 &  79.1\\
         {\sc LimeCal} & 69.3 & 82.9 & 72.0 & \bf 78.7 & 65.7 & 78.2 & 74.9 & 73.6 \\
         \bottomrule
    \end{tabular}
    \caption{Model performance and calibration performance of \limecal{} and glass box methods. On QA tasks, \limecal{} is better than {\sc Finetuning RoBERTa} and even outperforms {\sc Adapt Base QA/NLI} on \trivia{}. \limecal{} under-performs glass box methods on NLI due to its easy nature and the poor base-model performance.}
    \label{tab:finetune}
\end{table*}

\paragraph{Cross-Domain Generalization of Calibrators} Our calibrators so far are trained on individual transfer settings. Is the knowledge of a calibrator learned on some initial domain transfer setting, e.g., SQuAD $\rightarrow$ \trivia{}, generalizable to another transfer setting, e.g. $\rightarrow$ \hotpot{}? This would enable us to take our basic QA model and a calibrator and apply \emph{that pair} of models in a new domain without doing any new training or adaptation.
We explore this hypothesis on QA.\footnote{We also tested the hypothesis on the NLI-paraphrase transfer, but did not see evidence of transferability there, possibly due to the fact that these tasks fundamentally differ.}

For comparison, we also give the performance of a \roberta{}-model first finetuned on \squad{} and then finetuned on domain A ({\sc Adapt}, Figure~\ref{fig:setting}). {\sc Adapt} requires access to the model architecture and is an unfair comparison for other approaches.

We show the results in Table~\ref{tab:cross_qa}. None of the approaches generalize between \sqadv{} and the other domains (either trained or tested on \sqadv{}), which is unsurprising given the synthetic and very specific nature of \sqadv{}.

Between \trivia{} and \hotpot{}, both the \limecal{} and {\sc Kamath} calibrators trained on one domain can generalize to the other, even though \bowprop{} is not effective. Furthermore, our \limecal{} exhibits a stronger capability of generalization compared to {\sc Kamath}. We then compare \limecal{} against {\sc Adapt}. {\sc Adapt} does not always work well, which has also been discussed in \citet{selectiveqa, multiqa}. {\sc Adapt} leads to a huge drop in terms of performance when being trained on \hotpot{} and tested on \trivia{}, whereas \limecal{} is the best in this setting. From \trivia{} to \hotpot{}, {\sc Adapt} works well, but \lime{} is almost as effective.

Overall, the calibrator trained with explanations as features exhibits successful generalizability across the two realistic QA tasks. We believe this can be attributed to the features used in the explanation-based calibrator. Although the task is different, the calibrator can rely on some common rules to decide the reliability of a prediction.

\paragraph{Impacts of Training Data Size}

Calibrating a model for a new domain becomes cumbersome if large amounts of annotated data are necessary. We experiment with varying the amount of training data the calibrator is exposed to, with results shown in Table~\ref{tab:train_size}. Our explanation-based calibrator is still the best in every setting with as few as 100 examples. With 100 examples, {\sc Kamath} and \bowprop{} perform worse than the {\sc MaxProb} baseline on \trivia{} and \hotpot{}, indicating that more data is needed to learn to use their features. 

\subsection{Comparison to Finetuned Models}

Throughout this work, we have assumed a black box model that cannot be fine-tuned on a new domain. In this section, we compare calibration-based approaches with glass-box methods that require access to the model architectures and parameters. We evaluate two glass-box methods in two different settings (Figure~\ref{fig:setting}): (1) finetuning a base \roberta{} model  ({\sc Finetune RoBERTa}), which needs access to the model's architecture but not parameters; and (2) finetuning a base QA/NLI model, which requires both model architectures as well as parameters. All these models are finetuned with 500 examples, the same as \limecal{}. We also give the performance of a model trained with full in-domain training data for different tasks as references ({\sc InDomain QA/NLI}).

We present the model performance (measured with Exact Match and F1 for QA and Acc for NLI) and calibration results in Table~\ref{tab:finetune}. Note that there are no calibrators for glass box methods, so we only report AUC scores for calibration performance.

On QA tasks, the limited training data is not sufficient for successfully finetuning a \roberta{} model. Consequently, {\sc Finetune RoBERTa} does not achieve credible performance. Finetuning a base QA model greatly improves the performance, surpassing \limecal{} on \sqadv{} and \hotpot{}. However, we still find that on \trivia{}, \limecal{} slightly outperforms {\sc Adapt}. This is a surprising result, and shows that explanation-based calibrators can still be beneficial in some scenarios, even if we have full access to the model.

On NLI tasks that are substantially easier than QA, finetuning either a \roberta{} LM model or a base NLI model can reach an accuracy of roughly 80\%. Our explanation-based approach largely lags glass-box methods, likely because the base NLI model utterly fails on \qnli{} (50.5\% accuracy) and \mrpc{} (55.0\% accuracy) and does not grant much support for the two tasks. Nonetheless, the results on NLI still support our main hypothesis: explanations can be useful for calibration.


\begin{table}[t]
    \centering
    \small
    \begin{tabular}{cl|c|c|c}
            \toprule
         \multicolumn{2}{l|}{Known\, \textbackslash \, Unknown}  &  {\sc Sq-Adv} &  {\sc Trivia} & {\sc Hotpot}  \\
         \midrule
         \multirow{4}{*}{\STAB{\rotatebox[origin=c]{90}{\sc Sq-Adv}}}
         & {\sc MaxProb}  & 85.0 & 88.7 & 87.5 \\
         & {\sc Kamath} & 88.8 & 89.5 & 88.9 \\
         &{\sc BowProp} & 91.5 & 90.6 & 89.0 \\
         & {\sc LimeCal}& \bf 94.5 & \bf 91.7 & \bf 91.9 \\
        \midrule
         \multirow{4}{*}{\STAB{\rotatebox[origin=c]{90}{\sc Trivia}}}
         & {\sc MaxProb}  & 85.0 & 88.7 & 87.6 \\
         & {\sc Kamath} & 85.6 & 91.9 & 88.7 \\
         &{\sc BowProp} & 85.3 & 92.1 & 89.9 \\
         & {\sc LimeCal}& \bf  90.9 & \bf 92.5 & \bf 92.1 \\
        \midrule
         \multirow{4}{*}{\STAB{\rotatebox[origin=c]{90}{\sc Hotpot}}}
         & {\sc MaxProb} & 85.0 & 88.7 & 87.6 \\
         & {\sc Kamath} & 86.1 & 91.4 & 89.4 \\
         &{\sc BowProp} & 85.1 & 91.8 & 91.6 \\
         & {\sc LimeCal} & \bf 91.7 & \bf  92.3 & \bf 92.5 \\
         \bottomrule
    \end{tabular}
    \caption{Area under Coverage-F1 curve in the Selective QA setting. Our explanation-based approach is also strong in this setting, substantially outperforming existing baseline and our own ablation.}
    \label{tab:select_qa}
\end{table}

\section{Selective QA Setting}
\label{sec:selective}

Our results so far have shown that a calibrator can use explanations to help make binary judgments of correctness for a model running in a new domain. We now test our model on the selective QA setting from \citet{selectiveqa} (Figure~\ref{fig:setting}). This experiment allows us to more directly compare with prior work and see performance in a setting where in-domain (ID) and out-of-domain (OOD) examples are mixed together.

Given a QA model trained on source domain data, the goal of selective QA is to train a calibrator on a mixture of ID source data and \emph{known} OOD data, and test the calibrator to work well on a mixture of in-domain and an \emph{unknown} OOD data. 

We follow the similar experimental setup as in \citet{selectiveqa}. The detailed setting is included in the supplementary material.

\paragraph{Results}
As shown in Table~\ref{tab:select_qa}, similar to the main QA results. Our explanation-based approach, \limecal{}, is consistently the best among all settings. We point out our approach outperforms {\sc Kamath} especially in settings that involve \sqadv{} as known or unknown OOD distribution. This can be attributed the similarity between \squad{} and \sqadv{} which can not be well distinguished with features used in {\sc Kamath} (\texttt{\small Context Length, Answer Length}, and etc.). The strong performance of our explanation-based approach in the selective QA setting further verifies our assumption: explanation can be useful and effective for calibrating black box models.

\section{Related Work}

Our approach is inspired by recent work on the \emph{simulation} test \cite{doshi2017towards}, i.e., whether humans can simulate a model's prediction on an input example based on the explanations. Simulation tests have been carried out in various tasks \cite{anchor,nguyencomparing, chandrasekaran2018,evalai} and give positive results in some tasks \cite{evalai}. Our approach tries to mimic the process that humans would use to judge a model's prediction by combining heuristics with attributions instead of having humans actually do the task.

Using ``meta-features'' to judge a model also appears in literature on system combination
for tasks like machine translation \cite{bojar2017}, question answering \cite{selectiveqa,zhang2021}, constituency parsing \cite{charniak-2005, fossum-2009} and semantic parsing \cite{yin-2019-reranking}. The work of \citet{rajani-stacking} in VQA is most relevant to ours; they also use heuristic features, but we further conjoin heuristic with model attributions. Our meta-feature set is derived from the presence of certain properties, which is similar to the ``concepts'' used in concept-based explanations \cite{ghorbani2019towards,mu2020compositional},  but we focus on using them for estimating model performance rather than explaining a prediction.

Our work addresses the problem of calibration \cite{guocalib, desai2020}, which is frequently framed in terms of models' output probabilities. Past work has attempted to tackle this problem using temperature scaling \cite{guocalib} or label smoothing \cite{labelsmooth}, which adjust confidence scores for all predictions. In contrast, we approach this issue by applying a classifier leveraging instance-specific explanations. Past work on generalizing to out-of-domain distribution in NLP largely focuses on using unlabeled data from the target domain and requires finetuning a model \cite{madomain,adaptsurvey,guo2020multi}, whereas we improve OOD performance of strictly black-box models.

\section{Discussion \& Conclusion}
\paragraph{Limitations} Despite showing promising results in improving model generalization performance, our attribution-based approach does suffer from intensive computation cost. Using either \lime{} or \shap{} to generate attributions requires running inference a fair number of perturbations when the input size is large (see Appendix for details), which limits our method's applicability. But this doesn't undermine the main contribution of this paper, answering the question in the title, and our approach is still applicable as-is in the scenarios where we pay for access to the model but not per query.


\paragraph{Conclusion}
We have explored whether model attributions can be useful for calibrating black box models. The answer is \emph{yes}. By connecting attributions with human heuristics, we improve model generalization performance on new domains and tasks. Besides, it exhibits promising generalization performance in some settings (cross-domain generalization and Selective QA).

\section*{Acknowledgments}

Thanks to the anonymous reviewers for their helpful feedback.
This work was partially supported by NSF Grant IIS-1814522, a gift from Salesforce Inc, and a gift from Amazon.

\bibliography{anthology,custom}
\bibliographystyle{acl_natbib}

\appendix

\appendix
\onecolumn
\newpage
\twocolumn

\section{Details of the Kernel used in \lime{} and \shap{}}
\textbf{\textsc{Lime}} heuristically sets $\pi_x$ as an exponential kernel (with bandwith $\sigma$) defined on the cosine distance function between the perturbation and original input, i.e.,

\scriptsize
$$\pi_x(z)=\exp(-d_{\cos}(x,h_x(z))/\sigma^2)$$ 
\normalsize

\noindent That is, \lime{} assigns higher instance weights for perturbations that are closer to the original input, and so prioritizes classifying these correctly with the approximation. 

\textbf{\textsc{Shap}} derives the $\pi_x$ so the $\phi$ can be interpreted as Shapley values \cite{shapley1997value}:

\scriptsize
$$\pi_x(z)=\frac{n-1}{\binom{N}{|z|} |z| (n- |z|)}$$
\normalsize

\noindent where $|z|$ denotes the number of activated tokens (sum of $z$).
This kernel assigns high weights to perturbations with few or many active tokens, as the predictions when a few tokens' effects are isolated are important. This distinguishes \shap{} from \lime{}, since \lime{} will place very low weight on perturbations with few active tokens.

\section{Detailed Setup of Selective QA Setting}
We follow the similar experimental setup as in \citet{selectiveqa}. We train a \roberta{} QA model on \squad{}, and use on a mixture of 1,000 \squad{} dev examples + 1,000 known OOD examples to train the calibrator. We report test results on both the same type of mixture (1,000 \squad{} + 1,000 known OOD, diagonal blocks in Table~\ref{tab:select_qa}) and a mixture of 4000 \squad{} examples + 4,000 unknown OOD (2,560 \squad{} + 2,560 \sqadv{} as \sqadv{} only contains 2,560 examples).

\begin{table*}[t]
    \centering
    \small
    \begin{tabular}{lll|ll}
            \toprule
         {\sc Sq-Adv} & {\sc Trivia} & {\sc Hotpot} & {QNLI} & {\sc MRPC}\\
         \midrule
         {Attr to NNP in Q} &Prob of Top Pred  &  Prob of Top Pred & Attr Overlapping NN in H & Prob of Top Pred \\
         Attr to VB in C & Answer Length & Attr to Q & BOW Overl- NN in H & Attr to P\\
         Prob of Top Pred & Attr NNP in Q & Attr Wh- in Q& BOW Overl- NN in P & Attr to H\\
         Attr to NN in Q & Attr Wh- in Q & Attr to C& Attr to Non-Overl- NN in P & Attr to Non-Overl- NNP in H\\
         Answer Length & Attr to Question &  Attr to NNP in Q& Prob of Top Pred& Attr to Overl- SYM in P \\
         \bottomrule
    \end{tabular}
    \caption{Most important features used by the \limecal{} in different tasks. For QA, \texttt{\small Attribution of NNP in the question} and \texttt{\small Attribution of Wh- in the question} are generally important. For NLI, features related to overlapping/non-overlapping nouns are more effective.}
    \label{tab:feat_imp}
\end{table*}

\section{Feature Importance}
\label{sec:appendix}


We analyze the important features learned by the calibrator. We find explanation-based features are indeed generally among the top used features and more important than Bag-of-Word-based features (see the Appendix for a detailed list). All QA calibrators heavily rely on attribution values of the proper nouns (\texttt{\small NNP}) and wh-words in the question. BoW features of overlapping nouns are considered important on \qnli{}, but the top feature is still attribution-based.

These factors give insights into which parts of the QA or NLI reasoning processes are important for models to capture. E.g., the reliance on NNPs in \sqadv{} matches our intuitive understanding of this task: distractors typically have the wrong named entities in them, so if the model pays attention to NNPs on an example, it is more likely to be correct, and the calibrator can exploit this.

Table~\ref{tab:feat_imp} shows the most important features learned by \limecal{} for QA and NLI. For brevity, we present the features related to the probabilities of the top predictions into one feature (\texttt{Prob}). Explanation-based features are indeed generally among the top used features and more important than raw property features.

\section{Details of POS Tag Properties}
\label{sec:tag_detail}
We use tagger implemented in spaCy API.\footnote{\url{https://spacy.io/api}}
The tag set basically follows the Penn Treebank tag set except that we merge some related tags to reduce the number of features given the limited amount of training data.\footnote{\url{https://www.ling.upenn.edu/courses/Fall\textunderscore2003/ling001/penn_treebank_pos.html}}
Specifically, we merge  \texttt{\small JJ,JJR,JJS} into \texttt{\small JJ}, \texttt{\small NN,NNS} into \texttt{\small NN}, \texttt{\small NNP,NNPS} into \texttt{\small NNP},  \texttt{\small RB,RBR,RBS} into \texttt{\small RB}, \texttt{\small VB,VBD,VBG,VBN,VBP,VBZ} into \texttt{\small VB}, and \texttt{\small WDT,WP,WP\$,WRB} into \texttt{\small W}. In this way, we obtain a tag set of 25 tags in total.

 
\section{Details of Black Box Calibrators}


\paragraph{Feature Counts for QA}
\begin{itemize}
    \item {\sc Kamath} \cite{selectiveqa}: we use the \textbf{ 7} features described in \cite{selectiveqa}, including \texttt{\small Probability} for the top 5 predictions, \texttt{\small Context Length}, and \texttt{\small Predicted Answer Length}.
    
    \item \bowprop{}: In addition to the 7 features used in {\sc Kamath}. We construct the property space $\mathcal{V}$ as the union of low-level \texttt{\small Segment} and \texttt{\small Segment} $\times$ \texttt{\small Pos-Tags}. Since there are 3 segments  \texttt{\small question, context, answer} in the input, and 25 tags (Section~\ref{sec:tag_detail}), the size of the property space $|\mathcal{V}|$ is thereby given as $3 + 3 \times 25=78$. Therefore the total number of features (including the 7 from {\sc Kamath}) is {\bf 85}.

    \item \limecal{} and \shapcal{}: Recall that the size of the property space is 78. \limecal{} and \shapcal{} uses 78 features describing the attribution related to the corresponding properties in addition to the 85 features used in {\sc BowProp}. The total number of features is therefore  {\bf 163}.
\end{itemize}

\paragraph{Feature Counts for NLI}
\begin{itemize}
    \item {\sc ClsProbCal} \cite{selectiveqa}: we use \textbf{2} features in practice, \texttt{\small Probability of Entailment} and \texttt{\small Probability of Contradiction}. We do not include \texttt{\small Probability of Neutral} since it can be inferred from the probabilities of two other classes.
    
    \item \bowprop{}: In addition to the 2 features used in {\sc ClsProbCal}, we construct the property space $\mathcal{V}$ as the union of low-level \texttt{\small Segment} and \texttt{\small Segment} $\times$ \texttt{\small Pos-Tags} $\times$ \texttt{\small Overalapping Words}. Since there are 2 segments  (\texttt{\small Premise, Hypothesis}), 25 tags (Section~\ref{sec:tag_detail}), and 2 properties for overlapping  \texttt{\small Overlapping, Non-Overlapping}, the size of the property space $|\mathcal{V}|$ is given as $2 + 2 \times 25\times 2=102$. Therefore the total number of features (including the 2 from {\sc ClsProbCal}) is {\bf 104}.

    \item \limecal{} and \shapcal{}: \limecal{} and \shapcal{} add another 102 features in addition to the 104 features used in {\sc BowProp}. The total number of features are therefore  {\bf 206}.
\end{itemize}

\begin{table}[t]
    \centering
    \small
    \begin{tabular}{cl|cc}
            \toprule
         \multicolumn{2}{l|}{QA}  &  {\sc Num. Tree} &  {\sc Max Depth} \\
         \midrule
         \multirow{4}{*}{\STAB{\rotatebox[origin=c]{90}{\sc Sq-Adv}}}
         & {\sc Kamath} & 300 &  6 \\
         &{\sc BowProp} & 300 & 20 \\
         & {\sc LimeCal}& 300 & 20 \\
         & {\sc ShapCal}& 300 &  20 \\
        \midrule
         \multirow{4}{*}{\STAB{\rotatebox[origin=c]{90}{\sc Trivia}}}
         & {\sc Kamath} & 300 & 6 \\
         &{\sc BowProp} & 200 & 20 \\
         & {\sc LimeCal}& 300 & 20\\
         & {\sc ShapCal}& 300 &  20 \\
        \midrule
         \multirow{4}{*}{\STAB{\rotatebox[origin=c]{90}{\sc Hotpot}}}
         & {\sc Kamath} & 300 & 4 \\
         &{\sc BowProp} & 300 & 10 \\
         & {\sc LimeCal}& 300 & 10\\
         & {\sc ShapCal}& 300 & 10 \\
         \midrule
                 \multicolumn{2}{l|}{NLI}  &  {\sc Num. Tree} &  {\sc Max Depth} \\
         \midrule
         \multirow{4}{*}{\STAB{\rotatebox[origin=c]{90}{\sc QNLI}}}
         & {\sc Kamath} & 300 &  4 \\
         &{\sc BowProp} & 300 & 6 \\
         & {\sc LimeCal}& 400 & 20 \\
         & {\sc ShapCal}& 400 &  20 \\
        \midrule
         \multirow{4}{*}{\STAB{\rotatebox[origin=c]{90}{\sc MRPC}}}
         & {\sc Kamath} & 300 & 6 \\
         &{\sc BowProp} & 300 & 8 \\
         & {\sc LimeCal}& 400 & 20\\
         & {\sc ShapCal}& 400 &  20 \\

         \bottomrule
    \end{tabular}
    \caption{Hyperparameters used to train the random forest classifier for different approaches.}
    \label{tab:hyper_blackbox}
\end{table}

\paragraph{Cost of Generating Explanations} For QA tasks which have relatively long inputs, we sample 2048 perturbations and run inference over them for each example. For simpler NLI tasks, we use about 512 model queries for each example.

\paragraph{Hyperparameters}
We use the RandomForest implementation from Scikit-Learn \cite{JMLR:v12:pedregosa11a}.
We list the hyperparameters used in each approach in Table~\ref{tab:hyper_blackbox}. 
The hyperparameters are determined through grid search using 400 training examples and 100 validation examples. The choices of numbers of trees are [200, 300, 400, 500], and choices of max depth are [4, 6, 8, 10, 15, 20]. Then, for the experimental results in Table~\ref{tab:main_qa}, Table~\ref{tab:main_nli}, and Table~\ref{tab:train_size}, we always fix the hyper-parameters, and do not perform any further hyper-parameter tuning.

\section{Details of Glass Box Methods}
\paragraph{Finetuning RoBERTa}  For QA, we finetune the \roberta{}-base model with a learning rate of 1e-5 for 20 epochs (We also try finetuning for 3 epochs, but the objective does not converge with 500 examples.) We set the batch size to be 32, and warm-up ratio to be 0.06.

For MNLI, we finetune a \roberta{}-base model with a learning rate of 1e-5 for 10 epochs. We set the batch size to be 32, and warm-up ratio to be 0.06, following the hyper-parameters in \citet{roberta}.

\paragraph{Adapting a Base QA/NLI Model}  For QA, we adapt the base \roberta{} QA model trained on \squad{} with a learning rate of 1e-5 for 2 epochs. 

For MNLI, we finetune base \roberta{} NLI model trained on \mnli{} with a learning rate of 1e-5 for 10 epochs. Learning does not converge when finetuning for 2 epochs, as the \mnli{} task is too different from \qnli{} and \mrpc{}.


\end{document}